# Learning From What You Don't Observe

**Mark A. Peot**

Stanford University
Department of Engineering-Economic
Systems and Operations Research
Stanford, CA 94305
peot@enter.com

**Ross D. Shachter**

Stanford University
Department of Engineering-Economic
Systems and Operations Research
Stanford, CA 94305
shachter@stanford.edu

## Abstract

The process of diagnosis involves learning about the state of a system from various observations of symptoms or findings about the system. Sophisticated Bayesian (and other) algorithms have been developed to revise and maintain beliefs about the system as observations are made. Nonetheless, diagnostic models have tended to ignore some common sense reasoning exploited by human diagnosticians. In particular, one can learn from which observations have *not* been made, in the spirit of conversational implicature.

In order to extract information from the observations not made, we propose the following two concepts. First, some symptoms, if present, are more likely to be reported before others. Second, most human diagnosticians and expert systems are economical in their data-gathering, searching first where they are more likely to find symptoms present. Thus, there is a desirable bias toward reporting symptoms that are present. We develop a simple model for these concepts that can significantly improve diagnostic inference.

## 1 INTRODUCTION

"Is there any point to which you would wish to draw my attention?"

"To the curious incident of the dog in the night-time."

"The dog did nothing in the night-time."

"That was the curious incident," remarked Sherlock Holmes.

– from **Silver Blaze** by Sir Arthur Conan Doyle

This paper argues that the current practice for modeling missing observations in interactive Bayesian expert systems is *incorrect*. If there is no reported value for a chance node in a belief network, it is usually assumed that that chance node is unobserved and that this unobserved node contributes no likelihood information to the rest of the belief network. If this chance node has no graphical successors, then it is barren [Shachter, 86]: the joint distribution of the other variables in the belief network is not a function of the distribution of the unobserved node.

In interactive diagnostic expert systems, however, there is a systematic bias introduced by people's preferences for reporting that can lead to systematic errors in diagnosis. In medicine, common sources of reporting biases might include:

• a bias to report symptoms that are present instead of symptoms that are absent, or

• a bias to report symptoms that are more significant or urgent instead of symptoms that are less obviously urgent.

Failure to model these biases can lead an expert system to produce erroneous recommendations early in the diagnostic process. If these biases are explicitly modeled, we demonstrate that early diagnostic differentials are more focussed, leading to more focussed question asking. Furthermore, the observations that the user of the expert system chooses to reveal inform on the likelihoods of unobserved variables.

The primary objective of this paper is to describe a simple mechanism for modeling biases in the responses to *open probe* questions in diagnostic expert systems. Open probe questions are unstructured questions that give the user freedom to choose the order that information is revealed. Section 2 describes open and closed probe questions and demonstrates that the responses to open probe questions do influence the probabilities of unobserved symptoms in diagnostic belief nets. Section 3 introduces the *report node*–a mechanism for modeling open probe questions in diagnostic domains comprised of unrelated symptoms of comparable severity. Section 4 shows how the simple report mechanism can be modified to allow the expert system to infer a particular user's biases from their responses to open probe questions. Section 5 introduces a technique for handling symptoms that vary in perceived severity. Finally, Section 6 relates the report mechanism to concepts in linguistics, specifically *conversational implicature* and *scalar implicature*.

## 2 OPEN AND CLOSED PROBES

Questions in a diagnosis domain can be grouped into two general categories, *open* and *closed probes*. Closed probes are requests for specific items of information. Examples of closed probes include: "Is the patient male or female?" Or "Is the patient running a fever?" Closed probes tend to discourage the user of the expert system from volunteering more than just the requested information. An open probe is a more general request for information that allows the user to select from a plethora of possible responses. Examples of open probes might include: "What brings you to the doctor today?" or "What looks abnormal on the patient's CT scan?"

Open probe questions are more informative than closed probe questions even if the same observations are made. For example, suppose that a doctor starts a patient interview with the closed probe question: "Do you have a rash?" The response to this question ("Yes" or "No") will lead the doctor to adjust the probabilities of diseases that are conditionally relevant to the observation. The response will have no effect on the belief inferred for other diseases.

Say instead that the doctor starts the interview with the open probe question: "What is bothering you today?" The patient responds, "I have a rash." In this case, the doctor can conclude far more. In addition to the fact "Rash = true", the doctor might also reduce her belief that competing symptoms are present, especially if those symptoms are more severe. For example, the doctor might infer that it is unlikely that the patient is presenting with a severe headache or clenching chest pain; because if the patient were experiencing those symptoms, the patient would have mentioned them. This, in turn, decreases the probability that a non-rash-related disorder is present; focussing the doctor's subsequent question asking only on rash-related diseases.

Bayesian expert systems do not properly model the open probe phase of the patient interview process. In order to see why, we will consider the simple diagnostic belief network illustrated in Figure 1. In this figure, there are 4 binary chance nodes representing two competing diseases, *Poison Ivy* (PI) and *Migraine* (M), with their associated symptoms, *Rash* (R) and *Headache* (H). Each of these nodes is binary with values *absent* and *present*. *Poison ivy* causes *rash* and *migraine* causes *headache*.

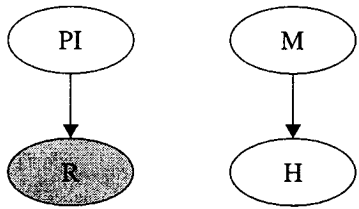

**Figure 1:** A simple diagnostic belief network. The shading on *R* denotes that the value for *R* is observed.

Suppose that we ask the patient a closed probe question about the presence of a rash. If the patient responds "Yes,"

we set the value of "Rash" in the belief network to "present." Based on this single observation, we conclude that the probability that poison ivy is present will change, while the probability that a migraine is present will be unchanged.

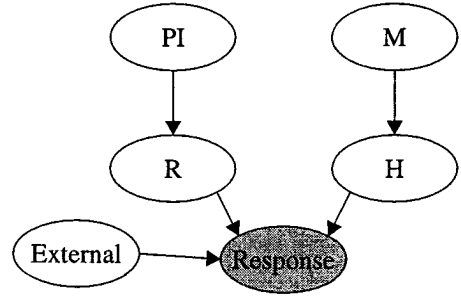

**Figure 2:** A belief network modeling a patient's response to an open probe question.

Now imagine that we ask the open probe question: "What is your problem?" If the patient responds that he has a rash and says nothing at all about the presence or absence of a headache, then we argue that the probability we should assign to headache (and migraine) should be reduced (or at least change).

The patient's response to the open probe question is a function of all of the symptoms $S_i$ that he is experiencing at the time that the question is asked as well as the history of interaction between the patient and the diagnostician and other external factors. This situation is illustrated in Figure 2. *Response* is a chance node that represents the patient's response to the open probe question. *Response* is a function of the presence or absence of each of the symptoms as well as the possibly-observable external factors node. The values for the *Response* variable include all of the possible responses that the patient might make when answering the open probe. Possibilities in this instance might include "I have a headache," or "I have a headache and a rash," "I hate it when you ask me all of these annoying questions," etc.

Because the patient's response to the open probe is a function of all of the symptoms that might influence the patient's chosen utterance, the response to the open probe should change our belief concerning every symptom that can influence the response. In this instance, we would argue that the patient would have a relatively high probability of reporting a severe headache if one were present. Since the patient did not choose to report any information about a headache, we can conclude that the patient probably does not have one.

### 2.1 THE IMPACT OF OPEN PROBE QUESTIONS ON EXPERT SYSTEMS

How does this problem manifest in a diagnostic expert system?

The architecture of a typical diagnostic expert system is illustrated in Figure 3. This expert system is based on the *hypothetico-deductive* (H-D) *cycle* [Gorry+Barnett, 68; Miller, et al, 82; Horvitz, et al, 84]. In the H-D cycle, the user initiates diagnosis by presenting a (possibly empty) set of salient observations to the expert system. These observations are fed into the expert system based on the implicit open probe, "What features describe your problem?" Based on these salient observations, the expert system computes a probability distribution $P(D_i)$ (called a *differential diagnosis*) over the possible fault or disease hypotheses $D_i$. After observing the initial differential diagnosis, the user can elect to act on this diagnosis or to refine the differential by answering further questions. In an expert system tool like Knowledge Industries' WIN-DX or Microsoft's MSBN system, some form of information value analysis [Howard 66; 67] is used to select the best observations to narrow the differential diagnosis. After the user answers one or more of these questions, the expert system formulates a new differential diagnosis and the cycle repeats.

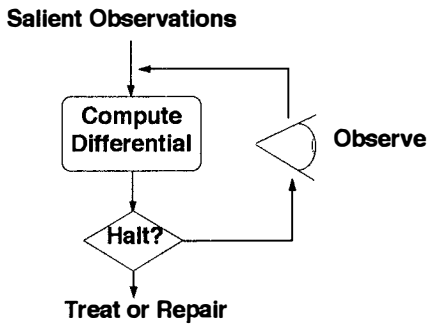

**Figure 3:** The H-D Cycle

In our example, the patient might enter the observation *Rash = present* in order to seed the diagnostic process. In the typical Bayesian expert system, a closed probe model (Fig. 1) is used to evaluate the response to the open probe question (Fig. 2). As a result, the expert system will over estimate the probability of symptoms the diseases that compete with *poison ivy* (e.g. *migraine*). This, in turn, causes the expert system to ask seemingly irrelevant questions in order to rule out competing diseases, even though no evidence has been presented that would lead the expert system to suspect that these diseases were present.

## 3 BASIC REPORT NODES

If we were *really* clever (Turing Award clever), we could assess the outcomes and probability distribution for the *Response* variable in Figure 2. We (unfortunately) are significantly less clever, so we propose just to roughly model the likelihood and independence properties of the portions of the belief network surrounding *Response*.

Figure 4 illustrates the likelihoods resulting from observing *Response* to be *"rash."* The report node mechanism (to be defined) is based on the following two assumptions about these likelihoods:

1. *Veridicality*: The user's response corresponds to the facts; that is, if the responder says that they have symptom *S*, then symptom *S* is present with probability **1.0**. This assumption allows us to assume that $\lambda^+$ (See Fig 4) is the same whether node *Rash* is observed directly or not.

2. *Likelihood Independence*: The likelihoods $\lambda^-(S_i)$ for symptoms that are not mentioned are mutually independent.[1]

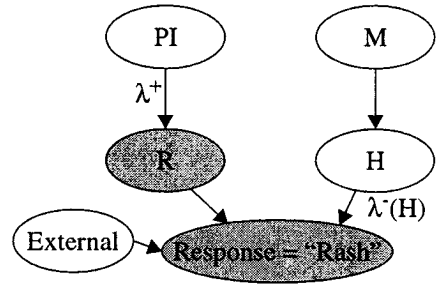

**Figure 4:** Likelihood feedback from *Response*.

Given these assumptions, we can derive a simpler network model (Figure 5) that has the same independence and likelihood properties as the network of Figure 4. We replace the *Response* node with a series of binary report nodes, $Report_Q(S_i)$; one for each symptom $S_i$ that is a predecessor of the old *Response* variable.

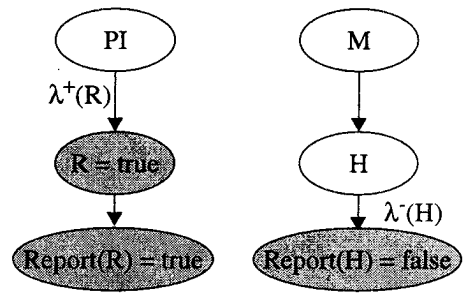

**Figure 5:** Approximating the response likelihood using *Report*.

The report node $Report_Q(S_i)$ represents the event: "A value is reported for symptom $S_i$ in response to an open probe question *Q*." $Report_Q(S_i)$ is *true* if any value is observed for $S_i$ and is *false* otherwise.

Note that if all of the symptom nodes were to be observed that the report nodes become independent from the remainder of the network. The report node mechanism is designed to capture the reporting biases for symptoms immediately after an open probe question is answered. Subsequent closed probe questioning on these unobserved symptoms will "wash away" the effect of the report nodes

---

[1] We will relax this assumption in the next section.

on those symptoms, D-separating [Geiger, et al; 90] the report nodes from the rest of the network.

## 3.1 ASSESSMENT: REPORTABILITY AND REPORTING BIAS

The probability distribution for $Report_Q(S_i)$ can be assessed directly. We have found it useful, however, to factor the assessment. First we aggregate symptom values into two categories: *present* and *absent*. For each symptom, we assess $P\{Report_Q(S_i) \mid S_i = present\}$, the probability that the symptom will be reported given that it is abnormal. We will call this probability the ***reportability*** of the symptom, $P_Q(S_i)$. Symptoms that are more evocative will tend to have a higher reportability. Subtle symptoms will tend to have a lower reportability.

The second parameter that we assess is the ***reporting bias*** for the symptom. This is the likelihood ratio

$$B_Q(S_i) = \frac{P\{Report_Q(S_i) \mid S_i = present\}}{P\{Report_Q(S_i) \mid S_i = absent\}} \quad .$$

The reporting bias captures the fact that people are more likely to report the presence of a symptom rather than the absence of that symptom (recall the quotation at the start of the paper). If the reporting bias is greater than or equal to 1, the reportability and reporting bias together specify a single unique conditional probability distribution for the report node.

In order to minimize the number of parameters that have to be assessed for a model, we can assume that the reporting bias is constant for all symptoms in the knowledge base. With this assumption, we only need to assess one parameter, reportability, for each symptom in the knowledge base. We can reduce the burden of assessment still further by assessing reportability for groups of symptoms rather than the individual symptoms themselves.

**Example:** Suppose that an expert system for dermatology proceeds sequentially through the following two phases of question-asking:

1. *Open probe (Initial complaint) phase*: The user selects any number of symptoms from a menu of complaints, pressing OK when done.

2. *Closed probe phase*: The diagnosis system uses information value analysis to select subsequent questions for the user to answer.

Our task is to assess a distribution for $Report_{Init}(Rash)$. *Rash* can be either a *bumpy blue rash*, an *itchy red rash*, or *absent*. In this case, the two types of rash are considered to have roughly the same reportability. The knowledge engineer feels that it is highly likely that the patient will report either of these rashes and assigns them a reportability of 0.95. The knowledge engineer had previously assigned a reporting bias of 5 to the entire knowledge base, indicating that the patient is 5 times more likely to report a symptom when the symptom is present.

The conditional probability distribution for $P\{Report_{Init}(Rash) \mid Rash\}$ is

|  | Rash = present | Rash= absent |
|---|---|---|
| $Report_{Init}(Rash)$ | $P_Q = 0.95$ | $\frac{P_Q}{B_Q} = 0.19$ |
| $\neg Report_{Init}(Rash)$ | $1 - P_Q = 0.05$ | $1 - \frac{P_Q}{B_Q} = 0.81$ |

## 3.2 REPORTABILITY, REPORTING BIAS AND INFERENCE

As mentioned earlier, observing symptom $S_i$ renders the report node $Report_Q(S_i)$ independent of the rest of the belief network. Thus, only the likelihood ratio

$$\frac{P\{\neg Report_Q(S_i) \mid S_i = present\}}{P\{\neg Report_Q(S_i) \mid S_i = absent\}} = \frac{B_Q(1 - P_Q)}{B_Q - P_Q}$$

has an effect on inference, providing a default likelihood function for each unobserved symptom. We call this likelihood function, the ***relative likelihood of no report*** or $\lambda_Q^-(S_i)$. As the reportability, $P_Q$, for $S_i$ approaches 1 (meaning that present symptoms are always reported), $\lambda_Q^-(S_i)$ tends to 0, implying that if symptom $S_i$ is not reported, it is almost certain that $S_i$ is absent. As the reportability for the symptom approaches 0, $\lambda_Q^-$ approaches 1; indicating that if there is only a low probability that $S_i$ will be reported if present, we should not assume that $S_i$ is absent if there is no report.

## 3.3 CUBITAL TUNNEL EXAMPLE

We tested the report node mechanism on the nerve compression disorder knowledge base (or *CTS*, after the most prominent disorder, carpal tunnel syndrome, in the KB). The CTS knowledge base was designed by a hand surgeon for evaluation of patients that were referred to his clinic with a preliminary diagnosis of a nerve compression injury affecting the hands or arms. Since the patients seen at this clinic have been referred by other doctors, the prior probabilities for many of the disorders are extremely large. For example, more than three-quarters of the patients in this clinic present with carpal tunnel syndrome.

We were interested in modeling the patient's response to an initial complaint question, such as "What is bothering you?" The response to this question can include any or all of the symptoms[2] in the knowledge base.

We modeled the bias in the response to this question by adding a report node successor to each of the symptoms that might reported in response to this question. In order to simplify the assessment process, we assumed a single

---

[2] As seen in Figure 6, there are four classes of observations: signs (observations made by the doctor), symptoms (observations made by the patient), test results and predisposing factors.

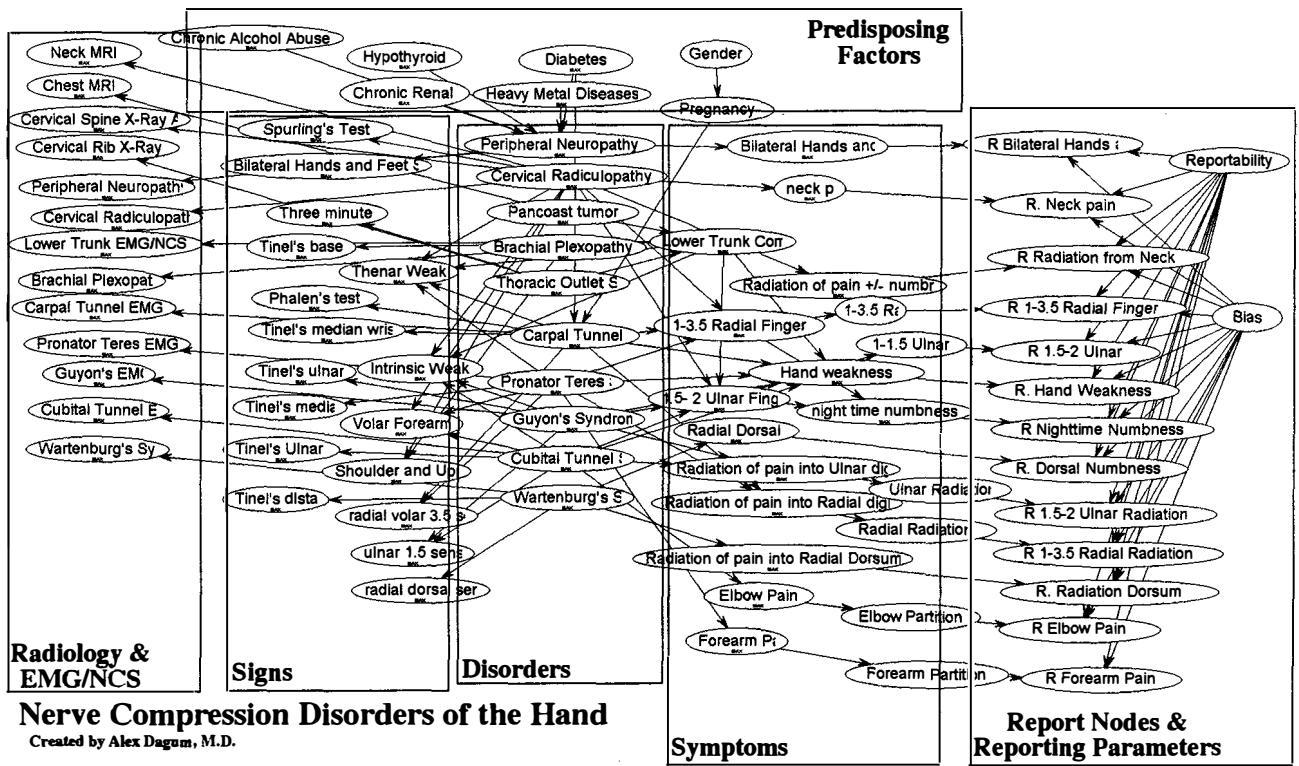

**Nerve Compression Disorders of the Hand**
Created by Alex Dagum, M.D.

**Figure 6:** Nerve Compression Disorders of the Hand

global value for the reporting bias, $B_{Global}$ and the reportability, $P_{Global}$, for all of the report nodes.

Figure 7 illustrates the effect of the reporting mechanism on the differential diagnosis as a function of $B_{Global}$. In this example, the reportability, $P_{Global}$, is set to 0.9. We entered three of the symptoms of cubital tunnel syndrome, a disorder caused by compression of the ulnar nerve where it passes through the cubital fossa (the "funny bone"). The symptoms:

1. Numbness or pain on both the volar (palm) and dorsal surfaces of the ulnar fingers (pinky and ring finger),

2. Radiation of pain into both the dorsal and volar surfaces of the ulnar fingers, and

3. Elbow pain with radiation of pain toward the hand.

Figure 7 was prepared by recording the probabilities for cubital tunnel syndrome and the next four leading disorders as a function of $B_{Global}$. If $B_{Global}$ is 1, the report nodes have no effect on the differential diagnosis. The probabilities for competing disorders are large because no symptoms have been observed that would rule these disorders out. In particular, note that carpal tunnel syndrome has nearly as high a probability as cubital tunnel syndrome, owing solely to its higher prior distribution.

As we increase the reporting bias, the probabilities for competing disorders drop by over an order of magnitude. The rate that the competing disorders drop in probability is different for each disease; probabilities for disorders with many obvious, but unobserved, symptoms will tend to be decrease more than the probabilities of disorders with less obvious symptoms. If only a subset of the 'classic' symptoms of a disorder are entered, the primary diagnosis will also drop in probability, however the drop will tend to be less than that of disorders that have no positive findings.

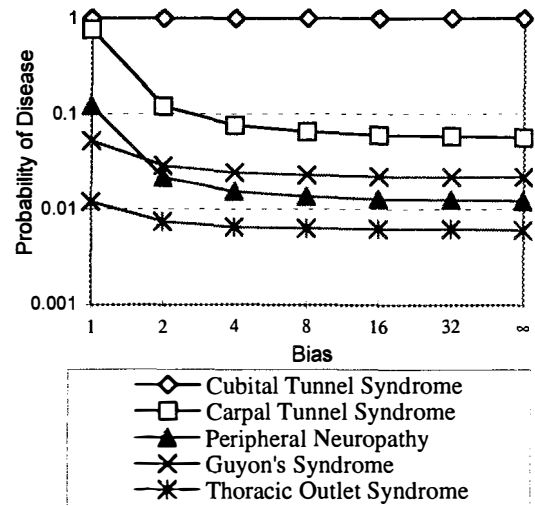

**Figure 7:** Differential diagnosis for the cubital tunnel syndrome case as a function of the global reporting bias $B_{Global}$.

Note that increasing bias focuses the initial diagnosis on those disorders that are relevant to the reported symptoms. Increasing either the report bias or the reportability further focuses the differential diagnosis.

## 4 LEARNING ABOUT REPORTABILITY

In the next two sections, we will describe ways to relax the likelihood independence assumption. In Section 5, we will model the interaction between the report nodes for symptoms of differing severities. In this section, we relate the likelihoods for the report nodes through uncertain global reportability and reporting bias variables.

The reporting preferences for individuals can be very different. For example, one patient might be more forthcoming about his symptoms, reporting the absence or presence of a multitude of symptoms. Another patient might be more reticent, reporting just a few abnormal symptoms. Can the reporting mechanism be applied to patients with widely varying reporting preferences?

As long as the patient obeys the veridicality assumption (e.g. he doesn't lie or indulge in strategic (gaming) behavior) and the reportabilities and biases for symptoms change proportionally, the answer is yes. It is relatively straight forward to adapt the report node mechanism to permit learning of the bias and reportability parameters for a patient, either based on the results of a single interview or multiple interviews.

The technique that we use is to condition the individual report nodes on chance nodes, $P_{Global}$ and $B_{Global}$, representing the global reportability and reporting bias for the knowledge base (see the far right side of Fig. 6). The reportability and bias for the individual report nodes can be any increasing function of these global reportability and reporting bias parameters. In the CTS knowledge base, we assume that the reportability and reporting bias for each of report nodes is the same as the global reportability.

When report nodes are instantiated, they update the densities for these global parameters. If a symptom is reported to be present or absent, the distribution over $P_{Global}$ is multiplied by the likelihood function $\lambda(P_{Global}) = P_{Global}$. If a symptom is not reported, the likelihood function used for updating $P_{Global}$ can't be factored from the rest of the network, but resembles a function of the form $\lambda(P_{Global}) = C - P_{Global}$ where $C \geq 1$ is a constant that depends on the probability that the symptom is actually *absent* and the value of the global bias. The likelihood functions for reportability roughly resemble those found in the conjugate beta distribution, however the distribution over reportability is not a conjugate distribution.

The likelihood update for the bias parameter $B_{Global}$ is a strong function of whether the symptom is *present* or *absent*. If mostly *present* symptoms are observed, probability for higher biases will increase (as expected). If more *absent* symptoms are observed, the expectation of the reporting bias decreases.

Figure 8 illustrates how the distribution over the global report bias and reportability changes with the number and type of observations in the CTS knowledge base. Since the reportability and bias variables cannot be represented using a conjugate distribution, we broke up the reportability and bias variables into a number of discrete values.[3] We assigned an equal prior probability to each of these values in order to enhance the effect of the report node likelihood functions. Each column in the chart illustrates the posterior belief for reportability and reporting bias as a function of the number and type of observations. All of the report nodes are instantiated in each of the cases–the value for each report node is set to true if its associated symptom is observed and is false otherwise.

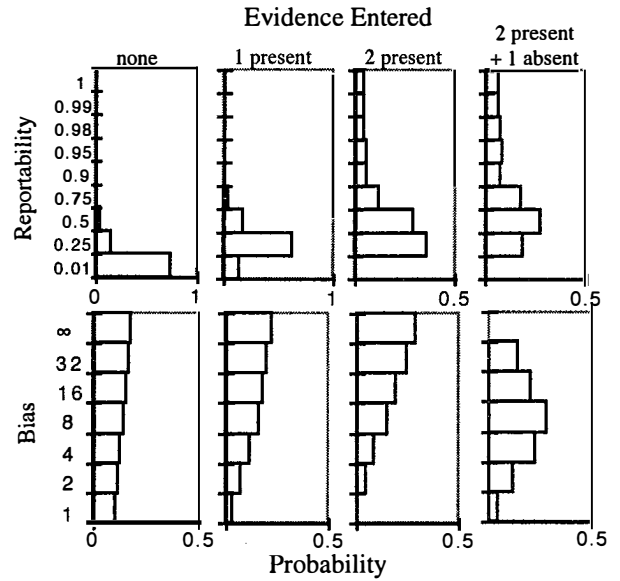

**Figure 8:** Learning Example

The far left column illustrates the belief in the reportability and report bias when no observations are entered in response to the initial complaint query. The expectation for the reportability is low (as expected). As more findings are observed, the expectation of reportability increases, regardless of whether the symptom reported is *present* or *absent*. Reportability, then, is primarily a function of the number of symptoms observed.

The reporting bias, on the other hand, is a strong function of whether the individual reports are present or absent. The middle columns of the figure illustrate how the expectation for the bias increases as more and more *present* symptoms are reported. The single report of *absent* (far right column) lowers the expectation of the bias.

## 5 REPORTING AND SEVERITY

In this section, we will consider the interaction between reports for symptoms of differing severities. If one symptom is perceived to be more severe than another, we might expect that the reportability for the first symptom is greater than that of the second. Furthermore, if a patient reports a number of mild symptoms at the start of an interview, we might conclude that it is extremely unlikely that a more severe symptom is present. In this section, we present a simple model that captures both of these kinds of inferences.

---

[3] An alternative approach might be to use an optimization routine to identify the setting for $P_{Global}$ and $B_{Global}$ that maximizes the probability of evidence for the entire network.

Throughout this section, we will consider only two classes of symptoms: *major* symptoms $\{S_1^*, ..., S_m^*\}$ and *minor* symptoms $\{S_1^-, ..., S_n^-\}$. Two uncertain global parameters, $P_{Major}$ and $P_{Minor}$, will be used to model the reportabilities for major and minor symptoms, respectively.

One possible belief network architecture that we might choose is to condition the reportability of the minor symptoms $P_{Minor}$ on the presence of (possibly unreported) major symptoms. For example, we can reducing the minor symptom reportability whenever the boolean expression $(S_1^* = present) \vee ... \vee (S_m^* = present)$ is true.

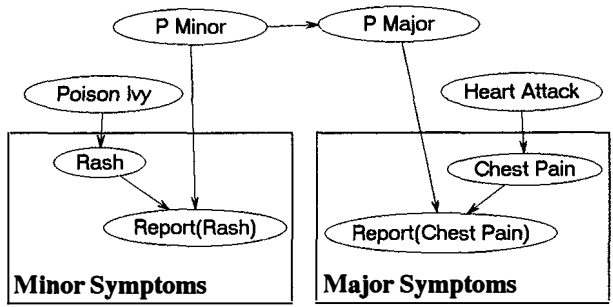

**Figure 9:** Reporting model for symptoms of differing severities.

We propose an alternative technique (Figure 9). The reportability for the major symptoms, $P_{Major}$, is a function $h(P_{Minor})$ of the reportability for the minor symptoms. This function satisfies the following properties.

1. *Increasing reportability*: $h$ is strictly increasing on $(0, 1)$. That is $h(P_{Minor}) > P_{Minor}$ when $P_{Minor} \in (0, 1)$.

2. *Probability*: $\forall (P_{Minor} \in [0, 1]), h(P_{Minor}) \in [0, 1]$

3. *Decreasing odds ratio*: $h(P_{Minor}) / P_{Minor}$ is decreasing.

It is easy to identify functions that satisfy these conditions. One such function is $h(x) = 2x - x^2$.

*Increasing reportability* ensures that the reportability of major symptoms is larger than the reportability of minor symptoms. *Decreasing odds ratio* guarantees the following: The change in expectation of reportability increases more when a minor symptom is reported to be present than if a major symptom is reported to be present. If this property is true, a report of a minor symptom will greatly increase the relative likelihood of no report for a major symptom, but the report of a major symptom will have a smaller effect on $\lambda_Q^-$. This means that if one or more minor symptoms is reported to be present and no major symptoms have been reported, then it is plausible to believe that no major symptoms are present. The converse is not true to the same degree.

For example, suppose that the prior probabilities for the disorders in Figure 9 are both 0.01. *Rash* and *Chest Pain* are both perfect observations–each symptom is present if and only if the corresponding disease is present. The prior distribution of $P_{Minor}$ is a uniform distribution and $P_{Major} = 2P_{Minor} - P_{Minor}^2$. If we observe *rash* to be *present* and *chest pain* is unreported, the probability of *heart attack* drops from a prior value of 0.01 to 0.0018. On the other hand, if we observe *chest pain* to be *present* and *rash* is unreported, the probability of *rash* drops from a prior of 0.01 to 0.004–a much smaller drop.

# 6 RELATED WORK

Our research on reportability seems strongly related to linguistics research on *conversational implicature*. Conversational implicature is based on the observation [Grice, 1975] that the participants in a conversation generally and cooperatively adhere to a set of conventions for generating contributions to that conversation. This *cooperative principle* allows the listener to infer more from a conversation than is literally expressed. Say that a speaker remarks "I have three children." This statement is literally true if the speaker has three, four, or twelve children. It is usually assumed, however, that if the speaker could have made a stronger claim ("I have twelve children."), he would have.

Grice's *Maxims of Conversation* [1967] represent an attempt (the most influential) to capture the conventions that govern the generation and interpretation of utterances in a conversation. Two of these maxims are relevant to the research reported in this paper, the *Maxim of Quality* and the *Maxim of Quantity*.

The Maxim of Quality says that a speaker should not say what he knows to be false and should not make any claims for which he lacks evidence. Veridicality (Section 3) is the assumption that the user of a diagnostic system will adhere to the Maxim of Quality.

Grice's Maxim of Quantity states that the speaker should make the strongest statement supported by the evidence available to him and not make a statement that is stronger. One classic example of this is the statement "Some of the apples are ripe." The Maxim of Quantity allows us to conclude from this statement that some of the apples are not ripe, because if all of the apples were ripe, the speaker would have said so. Our scheme for the interpretation of open probe questions relies on two quantity implicatures:

1. the assumption that a patient will report symptoms that are present more readily than symptoms that are absent, and

2. the assumpton that a speaker will tend to report more severe symptoms before those that are less severe.

This second implicature appears to be a kind of *scalar implicature* [Levinson, 83]. An *implicational scale* is a set of expressions of the same grammatical category that are linearly ordered in terms of their informativeness. For example, {all, many, some} constitute an implicational scale from the more informative "all" to the less informative "some." Now consider the statements "All of the apples are ripe," "Many of the apples are ripe," and "Some of the apples are ripe." Sentences using more informative terms from· an implicational scale tend to imply that weaker statements are true. For example, "Many of the apples are ripe" implies that "Some of the apples are ripe." Sentences using less informative statements tend to imply that stronger statements are false. For example, "Many of the apples are ripe" implies that the stronger statement "All of the apples are ripe" is false. In our work, symptoms of varying perceived severities appear to form a partial order similar to an implicational scale. A solitary report of a less severe symptom implies that a more severe symptom is absent. A solitary report of a severe symptom, on the other hand, does not seem to imply the absence of a lesser symptom quite as strongly.

# 7 CONCLUSIONS

We have presented a simple technique for capturing two kinds of reporting preferences for answers to open probe questions: The first is a preference for reporting present symptoms over absent symptoms. The second is a preference for reporting more severe symptoms before those that are less severe. These two techniques allow us to use the answers to open probes to infer not only the values for reported symptoms, but also to infer likelihood functions for unobserved symptoms. The likelihoods resulting from these unobserved symptoms has the effect of focussing the differential diagnosis, resulting in more focussed question asking. The technique is easy to apply, seems effective, and corrects a potentially serious source of bias in probabilistic expert systems.

It seems plausible that we might improve our model of open probe questions by exploiting contemporary research in linguistics. There is a question, though, about whether this is worth it. The primary effect of changing the open probe model is to change the observation order in the sequence of closed probe question-asking following the initial open probe. The final diagnosis may not be a strong function of the observation order. None-the-less, this seems to be a fruitful area for further experimentation and analysis.

## Acknowledgments

Alex Dagum (U. Toronto) developed the CTS knowledge base using Knowledge Industries' DXPress™ expert system tool. Eugene Charniak (Brown) first pointed out the link between reportability and conversational implicature. R. Michael Young (Pittsburg) and Barbara Grosz (Harvard) introduced us to scalar implicature. Denise Draper (Rockwell) made many useful presentation suggestions. We benefited greatly from the Summer Institute of Linguistics web-based glossary (www.sil.org) and Jeanne Frommer's web-based summary of *Pragmatics.*